\documentclass{article} 
\usepackage[utf8]{inputenc}
\usepackage{iclr2016_conference,times}
\usepackage{hyperref}
\usepackage{url}
\usepackage{scrextend} 
\usepackage{amsmath}
\usepackage{amssymb}
\usepackage{bm}

\title{An Exploration of Softmax Alternatives \\Belonging to the Spherical Loss Family}

\author{Alexandre de Brébisson and Pascal Vincent\thanks{and CIFAR} \\
MILA, Département d'Informatique et de Recherche Opérationnelle, \\University of Montréal\\
\texttt{alexandre.de.brebisson@umontreal.ca} \\
\texttt{vincentp@iro.umontreal.ca}
}

\iclrfinalcopy 

\begin{document}

\maketitle

\begin{abstract}

In a multi-class classification problem, it is standard to model the output of a neural network as a categorical distribution conditioned on the inputs. The output must therefore be positive and sum to one, which is traditionally enforced by a softmax. This probabilistic mapping allows to use the maximum likelihood principle, which leads to the well-known log-softmax loss. However the choice of the softmax function seems somehow arbitrary as there are many other possible normalizing functions. It is thus unclear why the log-softmax loss would perform better than other loss alternatives. In particular \cite{vincent2015efficient} recently introduced a class of loss functions, called the \emph{spherical family}, for which there exists an efficient algorithm to compute the updates of the output weights irrespective of the output size. In this paper, we explore several loss functions from this family as possible alternatives to the traditional log-softmax. In particular, we focus our investigation on spherical bounds of the log-softmax loss and on two spherical log-likelihood losses, namely the log-\emph{Spherical Softmax} suggested by~\cite{vincent2015efficient} and the log-\emph{Taylor Softmax} that we introduce. Although these alternatives do not yield as good results as the log-softmax loss on two language modeling tasks, they surprisingly outperform it in our experiments on MNIST and CIFAR10, suggesting that they might be relevant in a broad range of applications.

\end{abstract}

\section*{Introduction}

Classification problems with high dimensional outputs are particularly common in many language applications in which a target word has to be predicted out of a very large vocabulary. The standard application of backpropagation does not take advantage of the sparsity of the categorical targets and, as a result, the computations to update the weights of the output layer can be prohibitively expensive. Popular workarounds are based on approximations and can be divided into two main approaches. The first are sampling methods approximations, which compute only a tiny fraction of the output's dimensions (see for example \cite{Gutmann+Hyvarinen-2010, Mnih2013_2, Mikolov-et-al-NIPS2013, NIPS2014_5329}). The second is the hierarchical softmax, which modifies the original architecture by replacing the large output softmax by a heuristically defined hierarchical tree (\cite{Morin+al-2005, Mikolov-et-al-NIPS2013}).

\cite{vincent2015efficient} recently proposed an algorithm to compute the exact updates of the output weights in a very efficient fashion, provided that the loss belongs to a particular class of functions, which they call the spherical family because it includes an alternative to softmax, named \emph{spherical softmax} by \cite{Ollivier2013RiemannianMetricsArxiv}. In the rest of the paper, we call these losses the \emph{spherical losses}. If we denote $d$ the dimension of the last hidden layer and $D$ the dimension of the high dimensional output layer, they showed that for a spherical loss, it is possible to compute the exact updates of the output weights in $O(d^2)$ instead of the naive $O(d \times D)$ implementation. However it remains unclear how the spherical losses compare to the more traditional log-softmax loss in the context of classification. This is precisely what we aim to investigate in this paper.

We first describe precisely the spherical family and extract spherical bounds of the log-softmax loss from it. We then identify two particular normalizing activation functions, namely the \emph{spherical softmax} and the \emph{Taylor softmax}, that lead to log-likelihoods that belong to the spherical family and that may be suitable to train neural network classifiers. Finally we evaluate these different losses empirically by training models on several tasks: MNIST, CIFAR10/100 and language models on the Penntree bank and the One Billion Word dataset.

\section{Characterization of the spherical family}

Let $\bm{o}=W\bm{h}$ be the linear outputs of a neural network, where $\bm{o}$ has dimension $D$ and $\bm{h}$ represents the $d$ dimensional output of the last hidden layer. Let $\bm{y}$ be a sparse target and $\mathcal{A}(\bm{y})$ the indices of the non-zero elements of $\bm{y}$. The spherical family described in~\cite{vincent2015efficient} is composed of the functions that can be expressed using only the $o_{c}$ associated to non-zero $y_{c}$,  $q = \|\bm{o}\|^{2}=\sum_{i}o_{i}^{2}$ the squared
norm of the whole output vector and $s = \mathrm{sum}(\bm{o})=\sum_{i}o_{i}$:
\[spherical family = \mathcal{L}(s = \mathrm{sum}(o), q = \|o\|^{2}, \{(o_{c},y_{c}) | c\in\mathcal{A}(\bm{y})\}).\]

 \cite{vincent2015efficient} showed that for such loss functions, it is possible to compute the exact gradient updates of $W$ in $O(d^2)$ without even computing the output $\bm{o}$, instead of the $O(d \times D)$ naive implementation. As we focus on classification problems, we will assume for the rest of the paper that $\mathcal{A}(\bm{y})$ contains the single target class index $c$ corresponding to the single non-zero element $y_c$ of $\bm{y}$. The resulting family can be rewritten as follows:
 \[\mathcal{L}(s, q, o_{c}, y_{c}).\]

The square error following the linear output $\bm{o}$ belongs to this family:
\begin{align*}
L_{MSE}(o, y) &= \|o - y\|^{2} \\
&= q - 2o_{c}y_{c} + y_c^2.
\end{align*}
It is the loss of choice in regression problems. It is also sometimes used in classification problems\footnote{In classification, it is actually often used with a logistic sigmoid applied to $\bm{o}$ beforehand, but this results in a loss that does not belong to the spherical family} even though the log-softmax loss is nowadays considerably more popular. Contrary to the log-softmax loss, using the square error for classification does not correspond to the conditional likelihood of a categorical distribution. Nevertheless, like the log-softmax loss and other likelihood losses, the mean square error has the desirable property that its minimum is the conditional expectation.

\label{bounds}
\section{Spherical upper bounds of the log-softmax loss function}

In this Section, we consider functions from the spherical family that are upper bounds of the log-softmax loss:
\[
L(\bm{o}, c)=-\log\frac{e^{o_{c}}}{\sum_{k=1}^{D}e^{o_{k}}}=-o_c + \log\sum_{k=1}^{D}e^{o_{k}}.
\]

\cite{bouchard2007efficient} proposed the following upper bound for the log sum of exponentials (for any $\alpha\in\mathbb{R}$ and $\xi_{k}\in\mathbb{R}$):
\[
\log\sum_{k=1}^{D}e^{o_{k}} \le 
\alpha+\sum_{k=1}^{D}\frac{o_{k}-\alpha-\xi_{k}}{2}+\lambda(\xi_{k})((o_{k}-\alpha)^{2}-\xi_{k}^{2})+\log(1+e^{\xi_{k}}),
\]
where $\lambda(\xi)=\frac{1}{2\xi}(\frac{1}{1+e^{-\xi}}-\frac{1}{2})$. To be able to use the algorithm developed by \cite{vincent2015efficient}, the $\xi_{k}$ have to be the equal for all $k$. By replacing $\xi_{k}$ by $\xi$ and by optimizing $\alpha$ so that the bound is as tight as possible, we can derive the following\footnote{This derivation is a little tedious but trivial, we leave it out due to space constraints.} bound for $L(\bm{o}, c)$, which holds for any $\xi \in \mathbb{R}$:
\[
L \le \left(-\frac{(D-2)^{2}}{16D}\frac{1}{\lambda(\xi)}-\frac{D}{2}\xi-D\lambda(\xi)\xi^{2}+D\log(1+e^{\xi})+\frac{1}{D}s+\left(q-\frac{s^{2}}{D}\right)\lambda(\xi)-o_{c}\right).
\]

This bound clearly belongs to the spherical family. We tried two approaches to determine an optimal $\xi$: either considering it as a fixed hyperparameter or optimizing it for every example to yield the tightest bound. By minimizing this bound, we hope to minimize indirectly the negative log-softmax.

\label{spherical_likelihood}
\section{Spherical losses modeling categorical likelihoods}

In classification problems, it is standard to model the output as a categorical posterior distribution $P(categories|input)$. Hence, the computed output must consist of positive values that sum to one, which is generally enforced by a softmax function applied to the linear output $\bm{o}=W\bm{h}$. However, this property holds for a more general class of normalizing functions:

\begin{align*}
  \bm{f}_{norm} \colon \mathbb{R}^D &\to \mathbb{R}^D\\
  \bm{o} &\mapsto [f_{norm}(o)_k]_{1 \le k \le D},
\end{align*}
where
\[
\forall k, f_{norm}(\bm{o})_k = \frac{g_k(\bm{o})}{\sum_{i=1}^D g_i(\bm{o})},
\]
and where each $g_k$ has only positive values $g_k: \mathbb{R}^D \to \mathbb{R}^+$.

We can restrict this family to be component-wise:
\[\forall k: o \mapsto g_k(\bm{o}) = g(o_k), \]
with $g$ being a real function common to all the components.

Now that the output represents a categorical distribution,  the corresponding network can be trained by maximizing the likelihood on a training dataset, i.e. minimizing the negative log-likelihood (equivalently the cross-entropy)

\[ L_{log\_loss}(\bm{o}, c) = - \log \left( f_{norm}(\bm{o})_c \right), \]

where $c$ is the index of the target class for the example $\bm{o}$.


The exponential is commonly used for $g$, which gives the softmax function:
\[\bm{o} \mapsto f_{soft}(\bm{o})_k = \frac{\exp(o_k)}{\sum_{i=1}^D \exp(o_i)}. \]

However, despite being widely used, it remains unclear how the softmax compares to other normalizing functions. In particular, normalizing activation functions of the following form lead to log-likelihoods that belong to the spherical family:

\[ \bm{o} \mapsto f_{sph}(\bm{o})_k = \frac{a_1 + a_2 o_k + a_3 o_k^2}{\sum_k (a_1 + a_2 o_i + a_3 o_i^2)}, \]
where $a_1$, $a_2$ and $a_3$ are scalars such that $x \mapsto a_1 + a_2 x + a_3 x^2$ is a positive polynomial (which is equivalent to $a_3$ and $4a_1 a_3 - a_2^2$ being positive). The corresponding spherical log-likelihood loss can indeed be rewritten into the canonical form of the spherical family:

\begin{align*}
L_{log\_sph}(\bm{o}, c) &= - \log f_{sph}(\bm{o})_c\\
& = -\log \frac{a_1 + a_2 o_c + a_3 o_c^2}{a_1 D + a_2 s + a_3 q},
\end{align*}
In the next sections, we consider two particular instances of this family: the log-\emph{spherical softmax} and the log-\emph{taylor softmax}.

\subsection{Spherical softmax}

The first spherical alternative to the softmax function that we consider is the \emph{spherical softmax}, a minor modification of the  non-linearity investigated by \cite{Ollivier2013RiemannianMetricsArxiv} to which a small constant $\epsilon$ is added for numerical stability reasons:

\[ \bm{o} \mapsto f_{sph\_soft}(\bm{o})_k = \frac{o_k^2 + \epsilon}{\sum_{i=1}^D( o_i^2 + \epsilon)}. \]

The corresponding log-loss is the log-spherical softmax $L_{log\_sph\_soft}(\bm{o}, c) = -\log f_{sph\_soft}(\bm{o})_c$, whose gradient are 
\begin{align*}
\frac{\partial L}{\partial o_{c}}&=\frac{2o_{c}}{\sum_{i=1}^{D}(o_{i}^{2}+\epsilon)}-\frac{2o_{c}}{o_{c}^{2}+\epsilon},\\\frac{\partial L}{\partial o_{k\ne c}}&=\frac{2o_{k}}{\sum_{i=1}^{D}(o_{i}^{2}+\epsilon)},
\end{align*}
where $c$ is the index corresponding to the target class. From the expression of the gradients, we can see that $\epsilon$ is necessary to avoid numerical issues when either $\sum_{i=1}^D o_i^2$ or $o_c$ are very small. In practice, we found $\epsilon$ to be very important and it should be carefully tuned.

An interesting property of the spherical softmax (with $\epsilon = 0$) is that it is invariant to a global rescaling of the pre-activations $o$. This contrasts with the translation invariance of the softmax but it is unclear if this is a desirable property.

We can also notice that, contrary to the softmax function, the spherical softmax is even, i.e. it ignores the sign of the pre-activation $\bm{o}$. In the experiments Section, we will compare it to the softmax function taken on the absolute value of the pre-activations.

\subsection{Taylor softmax}

Our second spherical alternative to the softmax comes from the second-order Taylor expansion of the exponential around zero $\exp(x) \approx 1 + x + \frac{1}{2} x^2$, which leads to the following function, which we call the \emph{Taylor softmax}:

\[ \bm{o} \mapsto f_{tay\_soft}(\bm{o})_k = \frac{1 + o_k + \frac{1}{2} o_k^2}{\sum_{i=1}^D (1 + o_i + \frac{1}{2} o_i^2)}. \]

Its corresponding log-loss is the log-Taylor softmax
$L_{log\_tay\_soft}(\bm{o}, c) = -\log f_{tay\_soft}(\bm{o})_c$, whose gradient are 
\begin{align*}
\frac{\partial L}{\partial o_{c}}&=\frac{1+o_{c}}{\sum_{i=1}^{D}(1+o_{i}+\frac{1}{2}o_{i}^{2})}-\frac{1+o_{c}}{1+o_{c}+\frac{1}{2}o_{c}^{2}},\\
\frac{\partial L}{\partial o_{k\ne c}}&=\frac{1+o_{k}}{\sum_{i=1}^{D}(1+o_{i}+\frac{1}{2}o_{i}^{2})}.
\end{align*}
The numerator $1 + x_c + 0.5 x_c^2$ of the Taylor softmax is assured to be strictly positive and greater than $0.5$, its minimum value. The gradients are well-behaved as well, with no risk of numerical instability. Therefore, contrary to the spherical softmax, we do not need to use the extra hyperparameter $\epsilon$. Furthermore, unlike the spherical softmax, the Taylor softmax has a small asymmetry around zero.

\section{Experiments}

In this Section we compare the log-softmax and different spherical alternatives on several tasks: MNIST, CIFAR10/100 and a language modeling task on the Penntree bank and the One Billion Word datasets. Our goal was not to reach the state of the art on each task but to compare the influence of each loss. Therefore we restricted ourselves to reasonably sized standard architectures with little regularization, no ensembling and no data augmentation apart from CIFAR. In all the experiments, we used hidden layers with rectifiers, whose weights were initialized with a standard deviation of $\sqrt{\frac{2}{fan\_in}}$ as suggested in \cite{he2015delving}. For the output layers, we set the initial weights to zero. In our language experiments, we set the bias values such that the initial network outputs matched the prior frequencies of the classes.

We ran experiments to train neural language models with softmax outputs by minimizing the spherical upper bounds given in Section~\ref{bounds} but results were disappointing. Optimizing the bound actually degraded the initial perplexity (at initialization, the network outputs the frequencies of the words), which means that the minimum of the bound was worse than the simple initialization. In the subsections below, we will thus provide  detailed results only for the more promising spherical losses outlined in Section~\ref{spherical_likelihood}.
%
%


\subsection{MNIST}

We first compared the effectiveness of our different loss functions for training MNIST digit classifiers (\cite{LeCun98}). We used the same architecture for all the different losses: a convolutional neural network composed of two conv-pooling layers (30 and 60 feature maps, filter sizes 5, pooling windows of size 5) followed by a fully connected layer of 500 neurons and the output layer. We used rectifiers for all hidden neurons and initialized the weights with the He scheme (\cite{he2015delving}). The networks were trained with minibatches of size 200, a Nesterov momentum (\cite{sutskeverimportance}) of 0.9 and a decaying learning rate \footnote{we used the heuristic of dividing the learning rate by two every time the performance did not improve for 5 consecutive epochs.}. The initial learning rate is the only hyperparameter that we tuned individually for each loss. We used early stopping on the validation dataset as our stopping criterion.

\begin{table}[!ht]
\caption{Test set performances of a convolutional network trained on MNIST with different loss functions. For each loss, results were averaged over 100 runs (each with different splits of the training/valid/test sets and different initial parameter values), the standard deviation being in parenthesis. The loss column reports the training loss evaluated on the test set. negll refers to the negative log-likelihood. The \emph{log softmax abs} row corresponds to the log-softmax loss except that the softmax is applied on the absolute value of the pre-activations. The log-Taylor softmax outperforms the log-softmax, especially with respect to the negative log-likelihood.}
\label{tab:mnist}
\begin{center}
\begin{tabular}{| l || c | c | c |} 
 \hline 
 loss function & loss & error rate & number of epochs \\ [0.5ex] 
 \hline\hline
 MSE & mse: 0.0035 (0.00036) & 0.889\% (0.100) & 60 (17) \\ 
 \hline\hline
 Log softmax & negll: 0.0433 (0.0080) & 0.812\% (0.104) & 26 (7) \\ 
 \hline
 Log softmax abs & negll: 0.0437 (0.0097) & 0.813\% (0.095) & 25 (8) \\ 
 \hline
 Log spherical softmax & negll: 0.0311 (0.0031) & 0.828\% (0.094) &  27 (9) \\ 
 \hline
 Log Taylor softmax & negll: \textbf{0.0292} (0.0034) & \textbf{0.785\%} (0.097) &  22 (7) \\ 
 \hline
\end{tabular}
\end{center}
\end{table}

\begin{table}[!ht]
\caption{Test set performances of a convolutional network trained on MNIST with different loss functions trained and evaluated on the official training and testing sets of MNIST (contrary to the results of table~\ref{tab:mnist}, for which the data splits were random). For each loss, results were averaged over 100 runs with different initial random parameters, the standard deviation being in parenthesis. The loss column reports the training loss evaluated on the test set. negll refers to the negative log-likelihood. The results are significantly better than those reported in table~\ref{tab:mnist}, suggesting that the official MNIST set is particularly advantageous. The log-Taylor softmax still outperforms the log-softmax.}
\label{tab:mnist_orig}
\begin{center}
\begin{tabular}{| l || c | c | c |} 
 \hline 
 loss function & loss & error rate & number of epochs \\ [0.5ex] 
 \hline\hline
 Log softmax & negll: 0.0335 (0.0052) & 0.716\% (0.084) & 26 (7) \\ 
 \hline
 Log Taylor softmax & negll: \textbf{0.0247} (0.0020) & \textbf{0.688\%} (0.061) &  22 (8) \\ 
 \hline
\end{tabular}
\end{center}
\end{table}

In order to obtain results that more reliably reflect the effect of each individual loss, we repeated each training 100 times with different random splits of the training/validation/testing datasets and different initial random weight values. The results reported in table~\ref{tab:mnist} are the averaged scores obtained on the test set over all runs. The standard deviations are reported in parenthesis. Note that these results were computed and averaged on random splits of the training/valid/test datasets in order to be more reliable. We also trained the two best models on the original dataset split of MNIST and results are reported in table~\ref{tab:mnist_orig}: they are significantly better than those on random splits suggesting that the official testing set is simpler to classify than a randomly extracted one.

\subsection{CIFAR10}

CIFAR10 (\cite{KrizhevskyHinton2009}) is a dataset composed of 60k images of size $32 \times 32 \times 3$ and 10 output categories. For our experiments, we used a large convnet architecture of 14 layers with filters of size 3 and pooling windows of size 3 (inspired from the architecture of~\cite{Simonyan2015}). We used a weight decay, batch normalization~(\cite{Ioffe+Szegedy-2015}) and random horizontal flips. For the log softmax and the log Taylor softmax, we averaged the testing scores over 10 runs with different splits of the training/validation/testing datasets and different initial weight values. We tuned the initial learning rate for each loss function. Table~\ref{tab:cifar10} reports the performances on the test set with the different losses.

\begin{table}[!ht]
\caption{Test set performances of a convolutional network trained on CIFAR10 with different loss functions. For the log-softmax and the log-Taylor softmax, results were averaged over 10 experiments in order to be more reliable, the standard deviation being in parenthesis. For the MSE and the log-spherical softmax, we only had time to run a single experiment. The log-Taylor softmax outperforms the log-softmax.}
\label{tab:cifar10}
\begin{center}
\begin{tabular}{| l || c | c | c |} 
 \hline 
 Models & loss & error rate \\ [0.5ex] 
 \hline\hline
 MSE & mse: 0.0251 & 9.00\%\\ 
 \hline \hline
 Log softmax & negll: 0.411 (0.032) & 8.52\% (0.20) \\ 
 \hline
 Log spherical softmax & negll: 0.410 & 8.37\% \\ 
 \hline
 Log Taylor softmax & negll: \textbf{0.403} (0.034) & \textbf{8.07\%} (0.12)\\ 
 \hline
\end{tabular}
\vspace{1em}
\end{center}

\end{table}

\subsection{CIFAR100}

We used the same network and the same procedure as those of CIFAR10. We did not manage to train the network successfully with the MSE criterion (it yielded 99\% error rate). Table~\ref{tab:cifar100} reports the performances on the test set with the different losses.

\begin{table}[!ht]
\caption{ Performances on the test set of a convolutional network trained on CIFAR100 with different loss functions. For each loss, results were averaged over 5 experiments in order to be more reliable (the only difference being the initial random parameter values), the standard deviation being in parenthesis. The log-softmax outperforms the spherical losses.}
\label{tab:cifar100}
\begin{center}
\begin{tabular}{| l || c | c | c |} 
 \hline 
 Models & loss & error rate \\ [0.5ex] 
 \hline\hline
 Log softmax & negll: \textbf{1.69} (0.091) & \textbf{32.4\%} (0.85) \\ 
 \hline
 Log spherical softmax & negll: 1.90 (0.053) & 33.1\% (0.97) \\ 
 \hline
 Log Taylor softmax & negll: 1.88 (0.047) & 33.1\% (0.85)\\ 
 \hline
\end{tabular}
\vspace{1em}
\end{center}
\end{table}

\subsection{Language modeling}

\subsubsection{Penntree bank}

We trained word-level language models on the Penntree Bank (\cite{marcus1993building}), which is a corpus split into a training set of 929k words, a validation set of 73k words, and a test set of 82k words. The vocabulary has 10k words. We trained our neural language model (\cite{nnlm:2001:nips}) with vanilla stochastic gradient descent on mini-batches of size 250 using an input context of 6 words. For all the models, the embedding size is 250 and the hidden activation functions are rectifiers. For each loss function, we hyper-optimized the learning rate, the number of layers and the number of neurons per layer. For each model, we computed its perplexity on the test set, which is the exponential of the mean negative log-likelihood. We also computed the simlex-999 score (\cite{HillRK14}), which measures the quality of word embeddings based on the similarity between words as evaluated by humans. Table~\ref{tab:penntree} reports the results obtained by the best models for the different losses.

\begin{table}[!ht]
\caption{Comparison of different losses used to train a neural language model on the Penntree bank dataset. For each loss, the hyperparameters controlling the model architecture have been tuned individually to yield the best perplexity on the validation set. The top 10 error rate measures the proportion of time the target word is among the top 10 predicted words.}
\label{tab:penntree}
\begin{center}
\begin{tabular}{| l || c | c | c | c |} 
 \hline 
 Models and losses & Perplexity & top 10 error rate & Simlex 999 & number of epochs \\ [0.5ex] 
 \hline\hline
 [1] Log softmax & 126.7 & 0.501 & 0.109 & 6\\ 
 \hline
 [2] Log spherical softmax & 149.2 & 0.508 & 0.052 & 7\\ 
 \hline
 [3] Log Taylor softmax & 147.2 & 0.503 & 0.066 & 6\\ 
 \hline\hline

 [4] Log softmax abs & 128.2 & 0.503 & 0.0777 & 7\\ 
 \hline
\end{tabular}
\vspace{0.5em}
\end{center}

\begin{addmargin}[1.2em]{2em}
[1] two hidden layers of 809 neurons each.

[2] three hidden layers of 1264 neurons each. $\epsilon$ is set to 0.0198.

[3] three hidden layers of 1427 neurons each.

[4] Same architecture as [1] except that the softmax is applied to the absolute value of the pre-activations.
\end{addmargin}
\end{table}

\subsection{One billion word}

We also trained word-level neural language models on the One Billion Word dataset (\cite{ChelbaMSGBKR14}), which is composed of 0.8 billion words belonging to a vocabulary of 0.8 million words. For our experiments, we chose to restrict the vocabulary to 10k words. As this training dataset is almost 1000 times bigger than the Penntree bank dataset, we can not even do one full epoch and we are thus constantly in a regime of online learning in which each new training example has not been seen before. As a result, it is almost impossible to overfit the training dataset with reasonable size models. Bigger models tend to always perform better, so we chose to restrict ourselves to a few  architecture sizes and we compared the different losses on those, rather than doing an exhaustive architecture search for each case, as we did in the experiments on the PennTree bank dataset.

\begin{table}[!ht]
\caption{Comparison of the different losses used to train a neural language model on the One Billion Word dataset. The log-softmax outperforms the spherical losses, even though adding layers reduces the gap. For the log-softmax, adding hidden layers degrades the simlex score, while it improves it for the spherical losses.}
\label{tab:penntree}
\begin{center}
\begin{tabular}{| l || c | c | c |} 
 \hline 
 Models and losses & Perplexity & top 10 error rate & Simlex 999 \\ [0.5ex] 
 \hline\hline
 [0] Log softmax & 27.3 & 0.283 & 0.365\\ 
 \hline
 [0] Log Spherical softmax & 72.9 & 0.417 & 0.164\\
 \hline
 [0] Log Taylor softmax & 75.8 & 0.421 & 0.168\\ 
 \hline\hline
 [1] Log softmax & 19.4 & 0.245 & 0.336\\ 
 \hline
 [1] Log spherical softmax & 29.6 & 0.313 & 0.254\\ 
 \hline
 [1] Log Taylor softmax & 28.9 & 0.313 & 0.262\\ 
 \hline\hline
 [2] Log softmax & 19.2 & 0.244 & 0.318\\ 
 \hline
 [2] Log spherical softmax & 29.4 & 0.306 & 0.262\\ 
 \hline
 [2] Log Taylor softmax & 28.4 & 0.309 & 0.265\\ 
 \hline
\end{tabular}
\vspace{0.5em}
\end{center}

\begin{addmargin}[5em]{2em}
[0] no hidden layer

[1] one hidden layer of 1000 relus

[2] two hidden layers of 1000 relus
\end{addmargin}
\end{table}

\section{Discussion}

On MNIST and CIFAR10, the spherical losses work surprisingly well and, for the fixed architectures we used, they even outperform the log-softmax. This suggests that the log-softmax is not necessarily the best loss function for classification and that alternatives such as categorical log-losses from the spherical family might be preferred in a broad range of applications.


On the other hand, in our experiments with higher output dimensions, i.e. on CIFAR100, the Penntree bank and the one Billion Word dataset, we found that the log softmax yields better results than the log-spherical softmax and the log-Taylor softmax. The reasons for this apparent qualitative shift as the number of output categories increases remain unclear but we venture two hypothetical leads. The first is that the exponential non-linearity in the softmax boosts the large pre-activations relatively to the smaller ones a lot more than a squaring operation. It thus yields a stronger competition between pre-activations and a more discriminative behavior. It is possible that the resulting ability to more precisely single out a few top winning classes becomes increasingly crucial as the number of categories grows. Our second lead relies on the fact that the exponential of a linear combination of features is the \emph{product} of the exponentials of each weighted feature. Each of these exponential factors may represent a probability or an (unnormalized) density and the resulting product of these exponential factors can be seen as computing a conjunction (an  "AND") of features. This is not possible with the simple squared linear combinations of the spherical losses. 
 
%


To increase the flexibility of the networks with spherical losses, we tried to increase the non-linearity of the network prior to the loss by adding layers and by replacing the rectifiers by stronger nonlinearities. In particular, we tried to use a softmax as the activation function of our last hidden layer. We also tried to use directly the exponential as the activation function of the hidden layers. In this case, to avoid excessive values of the exponential, we used a truncated version of it and we also used batch normalization (\cite{Ioffe+Szegedy-2015}) to obtain reasonable ranges of pre-activations to prevent the exponential from exploding. Although we were able to train these models, they did not perform better than simple rectifier layers.

Among the approaches we explored, upper-bounding the negative log softmax with a spherical loss was an unsuccessful attempt. The spherical losses we tried were more promising and in particular the log-Taylor Softmax seems the most appropriate for classification. Contrary to the spherical softmax, it does not require the extra hyperparameter $\epsilon$, which can make the spherical softmax quite unstable and difficult to train. It also has a small asymmetry, which may be a desirable property.

\section*{Conclusion}

Our experiments showed that for several low dimensional problems, the log-softmax is surprisingly outperformed by certain losses of the spherical family, in particular the log-Taylor softmax. On the other hand, in higher dimensional problems, the log-softmax yields better results. The reasons of this qualitative shift remain unclear and further research should be carried out to understand it.

%
%

\subsubsection*{Acknowledgments}

We would like to thank Harm de Vries for helpful discussions about the optimization of the log spherical softmax and for providing us with a good baseline model for CIFAR10.

\bibliography{strings,strings-shorter,ml,aigaion-shorter,iclr2016_conference_v2}
\bibliographystyle{iclr2016_conference}

\end{document}